\documentclass{article}
\usepackage{ijcai21}

\usepackage{times}

\usepackage{soul}
\usepackage{url}
\usepackage[hidelinks]{hyperref}
\usepackage[utf8]{inputenc}
\usepackage[small]{caption}
\usepackage{graphicx}
\usepackage{amsmath}
\usepackage{booktabs}
\usepackage{multirow}
\usepackage{graphicx}
\usepackage[table,xcdraw,dvipsnames]{xcolor}
\usepackage[bordercolor=gray!0,backgroundcolor=red!10,linecolor=red!50,textsize=tiny,textwidth=0.56in,obeyFinal]{todonotes}
\usepackage{fontawesome}

\title{Learning User Embeddings from Temporal Social Media Data: A Survey}

\author{
Fatema Hasan$^1$\footnote{Contact Author}\and
Kevin S. Xu$^2$\and
James R. Foulds$^1$\And
Shimei Pan$^1$\\
\affiliations
$^1$Information Systems, University of Maryland, Baltimore County\\
$^2$Electrical Engineering \& Computer Science, University of Toledo\\

\emails
$^1$\{fhasan1, jfoulds, shimei\}@umbc.edu,
$^2$kevin.xu@utoledo.edu
}

\begin{document}

\maketitle

\begin{abstract}
User-generated data on social media contain rich information about who we are, what we like and how we make decisions. In this paper, we survey representative work on learning a concise latent user representation (a.k.a. user embedding) that can capture the main  characteristics of a social media user. The learned user embeddings can later be used to support different downstream user analysis tasks such as personality modeling, suicidal risk assessment and purchase decision prediction.  The temporal nature of user generated data on social media has largely been overlooked in much of the existing user embedding literature. In this survey, we focus on research  that  bridges the gap by incorporating temporal/sequential information in user representation learning.  
We categorize relevant papers along several key dimensions, identify limitations in the current work and suggest future research directions. 

\end{abstract}

\section{Introduction}
\noindent
The continual creation and archiving of rich user-generated content on social media sites has given us an opportunity to capture the main characteristics of social media users including their personality traits and decision-making processes. 
Past research has demonstrated that \textit{user embedding}, a mostly unsupervised machine learning process to derive concise latent user representations from raw social media content (e.g. text and image posts, likes, and friendship relations) is an effective approach for user modeling ~\cite{pan2019social,ding2017multi,pennacchiotti2011machine}. 
These learned user embeddings can later be used in diverse downstream user analysis tasks such as user preference prediction~\cite{pennacchiotti2011machine}, personality modeling~\cite{kosinski2013private}, substance use detection~\cite{ding2018interpreting},  social connection recommendation~\cite{liu2020dynamic} and depression detection~\cite{amir2017quantifying}.


An often overlooked aspect of social media based user embedding is the temporal nature of user-generated content, often spanning multiple years with precise timestamp annotations, which makes these data streams ideal for longitudinal analysis of human behavior. For example, by learning the social media posting patterns of individuals,  it is possible to track episodic outbursts of angry posts and make inferences about their personality, such as their level of impulsivity. 
%
%
Incorporating temporal information has traditionally been explored in the signal processing, ubiquitous computing, and network science research communities. There is also a large body of work in the natural language processing (NLP) community on sequential data mining. We argue that by exploring different technologies developed in diverse communities, we may uncover new solutions for temporal user embedding. 

Although smaller in number than papers on user embeddings that do not consider time, there has been a growing amount of research on temporal user embedding with social media data.  
Our survey is an attempt to summarize these emerging technologies and at the same time, provide the community some suggestions on how we can go forward. 

\begin{table*}[!htb]
\centering
\resizebox{.91\textwidth}{!}{%
\begin{tabular}{@{}lllll@{}}
\toprule\midrule
\multirow{2}{*}{\textbf{Paper}} &
  \multicolumn{1}{c}{\multirow{2}{*}{\textbf{\begin{tabular}[c]{@{}c@{}}Input Data\\ Type\end{tabular}}}} &
  \multicolumn{2}{l}{\textbf{Output  Embedding  Type}} &
  \multicolumn{1}{c}{\multirow{2}{*}{\textbf{Downstream Task}}} \\ \cmidrule(lr){3-4}\morecmidrules\cmidrule(lr){3-4}
 &
  \multicolumn{1}{c}{} &
  \textbf{\begin{tabular}[c]{@{}l@{}}Static/\\ Dynamic\end{tabular}} &
  \textbf{\begin{tabular}[c]{@{}l@{}}Single/\\ Joint\end{tabular}} &
  \multicolumn{1}{c}{} \\ \midrule\midrule
~\cite{sang2015probabilistic} &
  Text &
  Dynamic &
  Single &
  \begin{tabular}[c]{@{}l@{}}Personalized Information Recommendation, \\ Long-term and Short-term Interest\end{tabular} \\ \midrule
~\cite{liang2018dynamic} &
  Text &
  Dynamic &
  Joint (user-word) &
  \begin{tabular}[c]{@{}l@{}}Top-K Relevant and Diversified Keywords to \\ Profile Users' Dynamic Interests\end{tabular} \\ \midrule
~\cite{yin2014temporal} &
  User Activity &
  Dynamic &
  Joint (user-temporal) &
  Temporal Recommendation \\ \midrule
~\cite{khodadadi2018continuous} &
  User Activity &
  Static &
  Joint (user-temporal) &
  Time Prediction, Mark Prediction \\ \midrule
~\cite{yin2015dynamic} &
  User Activity &
  Dynamic &
  Joint (user-temporal) &
  Temporal Recommendation \\ \midrule
~\cite{li2017attributed} &
  Multimodal &
  Dynamic &
  Single &
  Node (User) Classification, Network Clustering \\ \midrule
~\cite{liu2020dynamic} &
  Multimodal &
  Dynamic &
  Single &
  \begin{tabular}[c]{@{}l@{}}Node (User) Classification, Network Reconstruction, \\ Link (Friendship) Prediction\end{tabular} \\ \midrule
~\cite{noorshams2020ties} &
  Multimodal &
  Static &
  Single &
  \begin{tabular}[c]{@{}l@{}}Fake Account Detection, Misinformation Detection, \\ ad Payment Risk Detection\end{tabular} \\ \midrule
~\cite{kumar2019predicting} &
  Multimodal &
  Dynamic &
  \begin{tabular}[c]{@{}l@{}}Joint (user-item\\ in separate space)\end{tabular} &
  \begin{tabular}[c]{@{}l@{}}Future Interaction Prediction, \\ User State Change Prediction\end{tabular} \\ \midrule
~\cite{fani2020user} &
  Multimodal &
  Static &
  Single &
  Personalized News Recommendation, User Prediction \\ \midrule
~\cite{costa2017modeling} &
  User activity &
  Static &
  Single &
  Bot Detection \\ \midrule
~\cite{yu2018netwalk} &
  Network &
  Static &
  Single &
  Anomaly Detection \\ \midrule
~\cite{xiong2019dyngraphgan} &
  Graph &
  Dynamic &
  Single &
  Link Prediction (Friendship), Link Reconstruction \\ \midrule
~\cite{gong2020exploring} &
  Network &
  Dynamic &
  Single &
  \begin{tabular}[c]{@{}l@{}}Link Reconstruction, \\ Changed Link Prediction\end{tabular} \\ \midrule
~\cite{beladev2020tdgraphembed} &
  Graph &
  Dynamic &
  Single &
  Anomaly Detection, Trend Analysis \\ \midrule
~\cite{gao2017novel} &
  Network &
  Static &
  Joint (user-info) &
  Information Diffusion Prediction \\ \midrule
~\cite{wu2017sequential} &
  Image &
  Static &
  Joint (user-photo) &
  Photo Popularity Prediction \\ \midrule
~\cite{wu2016unfolding} &
  Multimodal &
  Static &
  \begin{tabular}[c]{@{}l@{}}Joint (user-item\\ in separate space)\end{tabular} &
  Photo Popularity Prediction \\ \midrule
~\cite{yin2013unified} &
  Multimodal &
  Static &
  Joint (user-temporal) &
  Stable and Temporal Topic Detection \\ \midrule
~\cite{diao2012finding} &
  Text &
  Static &
  Joint (user-temporal) &
  Finding Bursty Topic \\ \midrule
~\cite{xie2016learning} &
  Multimodal &
  Dynamic &
  Joint (user-POI) &
  POI Recommendation \\ \midrule
~\cite{zhuo2019diffusiongan} &
  Multimodal &
  Static &
  Single &
  \begin{tabular}[c]{@{}l@{}}Information Diffusion Prediction, \\ Influence Relationship Prediction\end{tabular} \\ \midrule
~\cite{qiu2020temporal} &
  Multimodal &
  Static &
  Single &
  \begin{tabular}[c]{@{}l@{}}Link Prediction (Friendship), Node (User) Classification, \\ Network Reconstruction\end{tabular} \\ \midrule
~\cite{zhao2017geo} &
  User activity &
  Static &
  \begin{tabular}[c]{@{}l@{}}Joint (user-POI\\ in separate space)\end{tabular} &
  POI Recommendation \\ \midrule
~\cite{du2016recurrent} &
  User activity &
  Static &
  Joint (user-temporal) &
  Time Prediction, Event Prediction \\ \midrule
~\cite{yang2017decoupling} &
  Network &
  Static &
  Single &
  Temporal Link Prediction \\ \midrule
~\cite{xu2015stochastic} &
  Network &
  Dynamic &
  Single &
  Time Prediction \\ \midrule
~\cite{zhang2018user} &
  Multimodal &
  Static &
  Single &
  Photo Popularity Prediction \\ \midrule
~\cite{nguyen2018continuous} &
  Network &
  Static &
  Single &
  Temporal Link Prediction \\ \midrule\bottomrule
\end{tabular}%
}
\caption{High level summary of user embedding methods from temporal social media data}
\label{tab:overview}
\end{table*}

\section{Overview}
\noindent Given the limited space, we define 
a relatively narrow scope to include only \emph{user embedding methods that incorporate temporal information}.  Our scope includes node embedding methods that use dynamic social networks (where each \textit{node} represents a user). 
We exclude methods that do not use data from typical social media sites such as Facebook, Twitter, Flickr, Stack Overflow, Reddit, Google+, Digg, and Foursquare. We further refine the scope by including only methods published within the last ten years. Table~\ref{tab:overview} is an overview of these articles.  
We will occasionally discuss out-of-scope articles on  temporal modeling  methodology as they are applicable to social media data, but will not include them in the Table. 
We survey each article based on the input data type, output embedding type, and the downstream tasks that employ these embeddings.

\section{Input Data \& Output Embedding Type}

\noindent 
In the literature, we found five categories of input data being used to learn user embeddings from temporal social media streams: (i) text, (ii) image, (iii) user activity, (iv) network/graph, and (v) multi-modal.  \textit{Texts} may include  sequences of tweet streams or Facebook status updates, and \textit{images} are sequences of photos shared by users. User activity data refer to the timestamped records of users performing certain actions such as liking a post, rating a review, asking/answering a question (e.g., on Stack Overflow), 
and check-ins (e.g. on Facebook or Twitter). The dynamically created social networks (e.g. the friendship network on Facebook and the follower/retweet network on Twitter) constitute a sequence of networks at different timestamps. 
Finally, multi-modal data is a combination of multiple types of input data streams such as network-text, user activity-text, and image-user activity.

We also categorize output embeddings along two dimensions: (i) static \textit{vs}. dynamic, (ii) single \textit{vs}. joint-embedding.

\paragraph{Static \textit{vs.} Dynamic Embedding:} 
With a \emph{static} user embedding, for a given user, the system outputs only one  time-independent representation.
In contrast, \emph{dynamic} user embedding is a function of time,  which generates different user representations at different times. Hence, dynamic user embeddings can be a better option to model time-dependent aspects of user characteristics ~\cite{li2017attributed,liu2020dynamic}. 

\paragraph{Single \textit{vs.} Joint Embedding:} 
\emph{Single} user embedding only learns user representations. \emph{Joint} embedding, 
on the other hand, 
learns not only user representations but 
also representations of other related entities such as text, time, image, and item. Depending on whether different types of  embeddings are in the same or separate space, joint embedding can be divided into two categories: (i) \textit{shared} space, (ii) \textit{separate} space. For example,~\cite{liang2018dynamic} learn user and text representations jointly in the same \textit{shared} embedding space, while~\cite{kumar2019predicting} learn user and item representation jointly but they are in \textit{separate} embedding spaces. When user and entity embedding are in a shared space, in addition to user-user relations, we can also infer user-entity relations based on user and entity embeddings.

\vspace{0.1in}
In the following, we summarize the methodology of learning temporal user embeddings. As the temporal representation plays an important role in determining  proper temporal modeling methods, we first explain temporal representation.

\section{Temporal Representation}
\label{sec:temporal_representation}
There are various ways of encoding temporal information. 
We categorize them along two dimensions: (i) \emph{representation}: discrete \textit{vs.} continuous, (ii) \emph{duration}: timestamp \textit{vs.} interval-time \textit{vs.}  time-bin \textit{vs.} time-window  \textit{vs.} temporal order. 

\paragraph{Discrete \textit{vs.} Continuous}
In order to learn user representations with temporal data, we  need to decide how to represent time as a variable. With a \emph{discrete} representation, time is discretized into regularly spaced intervals~\cite{xu2015stochastic}. 
One may choose the granularity of these intervals (e.g. daily, monthly) based on the specific use case, sparsity of the events and the machine learning algorithms.
The \emph{continuous} representation of time, on the other hand, can accommodate irregularly spaced time intervals and allow the time variable to take the value of any real number. Therefore, within a time interval, in theory, there can be an infinite number of time points when an event can take place~\cite{yin2014temporal}

\paragraph{Timestamp \textit{vs.} Interval time \textit{vs.} Time-bin \textit{vs.} Time-window \textit{vs.} Temporal order} 
The input representation can also be classified in terms of duration. 
A \textit{timestamp} refers to the exact time of an action. For example, given a sequence of $n$ user posts created at different times $t$, we can record their timestamps as $(t_1, t_2, t_3, t_4, \ldots, t_i, \ldots, t_{n})$. Since a timestamp is represented as a real number, by definition, it employs a continuous time representation.  While timestamps can be directly used as an input to user representation learning algorithms, they can also be used for selecting input data in order to generate user embeddings at time $t$~\cite{li2017attributed}. 

\textit{Interval time} represents the time difference between two consecutive actions. For example, given a sequence of user posts at different times $(t_1, t_2, t_3, t_4, \ldots, t_i, \ldots, t_{n+1}) $, the interval time $\Delta_i$ can be defined as  $(\Delta_1, \Delta_2, \Delta_3, \ldots, \Delta_n) = (t_2 - t_1, t_3 - t_2, t_4 - t_3, \ldots, t_{n+1}-t_n)$.
Since the sequence of time intervals is irregularly spaced, this is a continuous time representation~\cite{yin2014temporal,noorshams2020ties}.

A number of articles~\cite{liang2018dynamic,wu2016unfolding,diao2012finding,xie2016learning,zhao2017geo} have also utilized a binning strategy to group input data in equal sized bins (e.g., weekly and monthly). \textit{Time-bins} can be considered as either a discrete or continuous time representation, depending on the relative resolution between the size of the bin and the time span of the entire dataset.  A \textit{daily} time bin may be considered a continuous representation if the entire dataset spans over 10 years.  It is typically considered discrete however if the time span of the entire dataset is only 10 days.

A \textit{time-window} is defined as a fixed context window to select input features~\cite{sang2015probabilistic}. A time-window can be used to generate embeddings at time $t$ with the input data in the $t-\Delta$ to $t+\Delta$ range, where $\Delta$ is a configurable time-window size variable.
Since it is regularly spaced, this is often considered a discrete-time representation. 

A \textit{temporal order} only preserves the sequential order of the inputs~\cite{zhang2018user}. A stream of tweets ordered by time falls into this category. Since it is not an explicit temporal representation, it is neither discrete nor continuous.

\begin{table*}[!htb]
\centering
\resizebox{\textwidth}{!}{%
\begin{tabular}{@{}lllllll@{}}
\toprule\midrule
\multirow{2}{*}{\textbf{Paper}} &
  \multicolumn{2}{c}{\textbf{Temporal Representations}} &
  \multicolumn{3}{c}{\textbf{ML Methologies}} &
  \multicolumn{1}{c}{\multirow{2}{*}{\textbf{Embedding Method}}} \\ \cmidrule(lr){2-3}\cmidrule(lr){4-6}\morecmidrules\cmidrule(lr){2-3}\cmidrule(lr){4-6}
 &
  \textbf{\begin{tabular}[c]{@{}l@{}}Continuous/\\ Discrete\end{tabular}} &
  \textbf{\begin{tabular}[c]{@{}l@{}}Interval time/\\ Timestamp/\\ Time-bin/\\ Time window\end{tabular}} &
  \textbf{\begin{tabular}[c]{@{}l@{}}Unsupervised/\\ Self-supervised/\\ Supervised\end{tabular}} &
  \textbf{\begin{tabular}[c]{@{}l@{}}Discriminative/\\ Generative/\\ Hybrid\end{tabular}} &
  \textbf{\begin{tabular}[c]{@{}l@{}}Matrix-factorization/\\ Probabilistic/\\ Neural Net/\\ Hybrid\end{tabular}} &
  \multicolumn{1}{c}{} \\ \midrule\midrule
~\cite{sang2015probabilistic} &
  Discrete &
  Time window &
  Unsupervised &
  Generative &
  Probabilistic &
  LDA \\ \midrule
~\cite{liang2018dynamic} &
  Discrete &
  Time-bin &
  Self-supervised &
  Hybrid &
  Hybrid &
  \begin{tabular}[c]{@{}l@{}}Skip-gram extended\\ by Kalman filter\end{tabular} \\ \midrule
~\cite{yin2014temporal} &
  Continuous &
  Interval time &
  Unsupervised &
  Generative &
  Probabilistic &
  Topic Model \\ \midrule
~\cite{khodadadi2018continuous} &
  Continuous &
  Timestamp &
  Unsupervised &
  Generative &
  Probabilistic &
  Temporal Point Process \\ \midrule
~\cite{yin2015dynamic} &
  Discrete &
  Time window &
  Unsupervised &
  Generative &
  Probabilistic &
  Topic Model \\ \midrule
~\cite{li2017attributed} &
  Continuous &
  Timestamp &
  Unsupervised &
  N/A &
  Matrix-factorization &
  Matrix Decomposition \\ \midrule
~\cite{liu2020dynamic} &
  Continuous &
  Timestamp &
  Unsupervised &
  N/A &
  Matrix-factorization &
  Matrix Decomposition \\ \midrule
~\cite{noorshams2020ties} &
  Continuous &
  Interval time &
  Supervised &
  Discriminative &
  Neural Net &
  Attention based Encoder \\ \midrule
~\cite{kumar2019predicting} &
  Continuous &
  Timestamp &
  Self-supervised &
  Discriminative &
  Neural Net &
  RNN, Attention \\ \midrule
~\cite{fani2020user} &
  Continuous &
  Interval time &
  Unsupervised &
  Discriminative &
  Neural Net &
  Graph Embedding \\ \midrule
~\cite{costa2017modeling} &
  Continuous &
  Timestamp &
  Unsupervised &
  Generative &
  Probabilistic &
  Stochastic Process \\ \midrule
~\cite{yu2018netwalk} &
  Continuous &
  Timestamp &
  Unsupervised &
  Discriminative &
  Neural Net &
  \begin{tabular}[c]{@{}l@{}}Graph Embedding, \\ Autoencoder\end{tabular} \\ \midrule
~\cite{xiong2019dyngraphgan} &
  Continuous &
  Timestamp &
  Unsupervised &
  Generative &
  Neural Net &
  GAN \\ \midrule
~\cite{gong2020exploring} &
  Discrete &
  Time window &
  Unsupervised &
  Discriminative &
  Neural Net &
  GCN, LSTM \\ \midrule
~\cite{beladev2020tdgraphembed} &
  Discrete &
  Time window &
  Self-supervised &
  Discriminative &
  Neural Net &
  \begin{tabular}[c]{@{}l@{}}Graph Embedding \\ extended by CBOW\end{tabular} \\ \midrule
~\cite{gao2017novel} &
  Continuous &
  Temporal order &
  Self-supervised &
  Discriminative &
  Neural Net &
  Word2Vec \\ \midrule
~\cite{wu2017sequential} &
  Continuous &
  Time-bin &
  Supervised &
  Discriminative &
  Neural Net &
  CNN, LSTM \\ \midrule
~\cite{wu2016unfolding} &
  Discrete &
  Time-bin &
  Supervised &
  N/A &
  Matrix-factorization &
  Matrix Decomposition \\ \midrule
~\cite{yin2013unified} &
  Discrete &
  Time window &
  Unsupervised &
  Generative &
  Probabilistic &
  LDA \\ \midrule
~\cite{diao2012finding} &
  Discrete &
  Time-bin &
  Unsupervised &
  Generative &
  Probabilistic &
  LDA \\ \midrule
~\cite{xie2016learning} &
  Discrete &
  Time-bin &
  Unsupervised &
  Generative &
  Probabilistic &
  Graph Embedding \\ \midrule
~\cite{zhuo2019diffusiongan} &
  Continuous &
  Temporal order &
  Supervised &
  Generative &
  Neural Net &
  GAN \\ \midrule
~\cite{qiu2020temporal} &
  Continuous &
  Timestamp &
  Supervised &
  Hybrid &
  Neural Net &
  \begin{tabular}[c]{@{}l@{}}Temporal Random Walk, \\ Autoencoder\end{tabular} \\ \midrule
~\cite{zhao2017geo} &
  Discrete &
  Time-bin &
  Self-supervised &
  Discriminative &
  Neural Net &
  Skip-Gram \\ \midrule
~\cite{du2016recurrent} &
  Continous &
  Timestamp &
  Supervised &
  Hybrid &
  Hybrid &
  \begin{tabular}[c]{@{}l@{}}RNN, Temporal \\ Point Process\end{tabular} \\ \midrule
~\cite{yang2017decoupling} &
  Continuous &
  Timestamp &
  Unsupervised &
  Generative &
  Probabilistic &
  Temporal Point Process \\ \midrule
~\cite{xu2015stochastic} &
  Discrete &
  Time-bin &
  Unsupervised &
  Generative &
  Probabilistic &
  Autoregressive HMM \\ \midrule
~\cite{zhang2018user} &
  N/A &
  Temporal order &
  Supervised &
  Discriminative &
  Neural Net &
  \begin{tabular}[c]{@{}l@{}}VGGNet, LSTM, \\ Hierarchical Attention\end{tabular} \\ \midrule
~\cite{nguyen2018continuous} &
  Continuous &
  Timestamp &
  Self-supervised &
  Discriminative &
  Neural Net &
  Temporal Random Walk \\ \midrule\bottomrule
\end{tabular}%
}
\caption{Temporal representations and methodologies used in generating user embedding from temporal social media data}
\label{tab:methodologies}
\end{table*}

\section{Methodology}
\label{sec:methods}
In this section, we summarize the main methods for temporal user embeddings.  We categorize them first in terms of machine learning methodology (shown in Table~\ref{tab:methodologies}) and second in terms of temporal modeling methodology. 

\subsection{Machine Learning (ML) Methodology}
The papers in our survey cover a wide variety of techniques originated from diverse fields such as statistical learning theory, probabilistic modeling, graph/network theory, and neural networks; they also vary widely in the amount of supervision they receive from labeled ground truth data, as well as their 
methods of modeling the input data space. Here, we categorize the machine learning methodology used in these papers along three dimensions:

\paragraph{Unsupervised \textit{vs.} Self-supervised \textit{vs.} Supervised}

The main difference between unsupervised and supervised ML pertains to the use of ground truth labels; the unsupervised approaches do not require any ground truth labels to derive  embedding features while supervised methods require ground truth labels from the target task. 

A self-supervised approach is a special case of supervised ML where training examples can be automatically constructed from raw input data (a.k.a.~no human-provided ground truth labels are required). 
For this reason, self-supervised approaches are also sometimes referred to as unsupervised. 
Self-supervised approaches often rely on  \textit{auxiliary} training tasks for which a large number of training instances can be automatically constructed. 
\texttt{Word2Vec}~\cite{mikolov2013distributed} is a classic example of a self-supervised approach, learning the representation of words by predicting other words in its context (\texttt{Skip-Gram}). A user representation can be derived by aggregating all the words authored  by the same user within a temporal context. 

Supervised learning can be used to learn temporal user embeddings ~\cite{zhuo2019diffusiongan,qiu2020temporal,wu2016unfolding,wu2017sequential,noorshams2020ties}, although this is uncommon. Unlike unsupervised or self-supervised embeddings, supervised user embeddings may not be generalizable as they are optimized specifically for only the target task. 

In terms of the specific embedding methods employed, Latent Dirichlet Allocation (LDA)~\cite{blei2003latent} is one of the most popular unsupervised feature learning methods employed in the survey articles to characterize a person using a mixture of topics conveyed in their social media posts~\cite{sang2015probabilistic,yin2014temporal,yin2015dynamic,yin2013unified,diao2012finding}. Various extensions to LDA (e.g., Dynamic LDA~\cite{blei2006dynamic}) have been proposed to  model topic distributions across time. Other notable unsupervised methods include
Temporal Point Process~\cite{costa2017modeling,khodadadi2018continuous},
Spectral Embedding of Graphs~\cite{li2017attributed}, and
Singular Value Decomposition~\cite{liu2020dynamic}. Recently, most modern user embeddings approaches have used neural network-based self-supervised learning~\cite{kumar2019predicting,beladev2020tdgraphembed} due to the flexibility and scalability of these approaches. 

\paragraph{Discriminative \textit{vs.} Generative}
The main difference between generative and discriminative models boils down to how they model the input data distribution from a probabilistic perspective. 
Generative models learn the joint probability distributions of input and output, whereas discriminative models learn the conditional probability distribution of the output, given the input.
Due to the generative model's ability to model the distribution of the input data itself (as opposed to learning the mappings between the input and output),
a large portion of the unsupervised methods follow a generative methodology such as LDA~\cite{sang2015probabilistic,yin2015dynamic} and
Temporal Point Process~\cite{costa2017modeling,khodadadi2018continuous}.

Combining adversarial learning with generative models (as in GAN) has been gaining traction in modeling high-order proximity and temporal evolution in graph-structured data~\cite{xiong2019dyngraphgan} as these two properties were found to be particularly challenging to learn through a purely generative process~\cite{zhou2018dynamic}. Under the GAN framework, a \textit{generator} is tasked with creating a graph-based representation of the networks' temporal evolution and the \textit{discriminator} is used to assess the probability of the generated representation being real. Such models can be efficiently trained with back-propagation without relying on costly sampling strategy.

Dynamic \textit{graph embeddings} can also be learned in a purely discriminative manner as long as sufficient ground truth for the training task is  available~\cite{noorshams2020ties,fani2020user,yu2018netwalk,beladev2020tdgraphembed}. Other popular discriminative architectures include \texttt{Word2Vec}~\cite{gao2017novel,zhao2017geo} and autoregressive models (e.g. recurrent architectures)~\cite{kumar2019predicting,gong2020exploring,wu2017sequential}.

Last but not the least, several works have also had success with using a mixture of both generative and discriminative methodologies.  For example, \cite{liang2018dynamic} extended a (discriminative) \texttt{Skip-Gram} model 
with a (generative) Kalman filter to capture the temporal dynamics of Twitter profiles. Similarly, ~\cite{qiu2020temporal} combined a temporal random walk method with a supervised auto-encoding architecture to learn temporal network embeddings.

\paragraph{Matrix-factorization \textit{vs.} Probabilistic \textit{vs.} Neural Network}
\noindent
Among the 29 articles we surveyed, around half of them use neural-networks as the underlying embedding-learning method. Others mostly follow a probabilistic approach, with only 3 utilizing matrix-factorization. Once a staple in text-based embedding learning and recommender systems, matrix decomposition techniques have fallen out of favor due to the prevalence of probabilistic (e.g. LDA) and neural-network based (e.g. \texttt{Word2Vec}) embedding methods. However, they are still used for constructing low dimensional representation of complex dynamic networks  where the adjacency matrix remains incomplete and noisy~\cite{li2017attributed,liu2020dynamic,wu2016unfolding}. 

Among the probabilistic approaches, the works of~\cite{diao2012finding,yin2013unified,yin2014temporal,yin2015dynamic,sang2015probabilistic} are notable for extending LDA across time to create temporal topic model. However, for more specialized use cases (e.g. predicting specific user action in time) we identified the usage of statistical processes such as Temporal Point Processes~\cite{khodadadi2018continuous,costa2017modeling,yang2017decoupling}, which we discuss further in Section \ref{sec:temporalMethod}. 

Autoregressive models such as Recurrent Neural Net (RNN) and Long Short Term Memory (LSTM) are appropriate options for modeling time series data within a neural-network framework; and have found success in learning temporal user embeddings from graph-structured data~\cite{kumar2019predicting,gong2020exploring,wu2017sequential}. 
Here the input is a sequence of text/image/user activity
selected under a specific temporal context (e.g. temporal neighborhood or periodic neighborhood).
The learned user embeddings are often represented by the sequence of the hidden states, which are fed as the input to the downstream applications.
It is worth noting that each of these approaches required specialized feature extractors (e.g. attention mechanism or graph convolution) to learn spatial (e.g. graph node vicinity) or short-range temporal dependency before the long-range temporal dependency can be learned by the following recurrent network layers~\cite{gong2020exploring}.

\subsection{Temporal Modeling Methodology}
\label{sec:temporalMethod}
\noindent 
A large portion of the articles directly extends a well established method (e.g. LDA or \texttt{Word2Vec}) with a temporal aspect.
Only a small portion of the literature has developed dedicated techniques to extract the temporal characteristics of data (e.g.~temporal point processes~\cite{khodadadi2018continuous} and RNN/LSTM~\cite{kumar2019predicting,gong2020exploring}).
In the following, we discuss these temporal modeling techniques in  details. 

\paragraph{Probabilistic models for discrete time}
A common approach to extend a model designed for a static setting, e.g.~LDA, to one that evolves in discrete time bins is to assume a hidden Markov model (HMM) structure. 
HMM models the observed data in each time bin using the static model but chains together the unobserved or latent parameters for different time bins using a Markov model. 
These models have been used to chain parameters in discrete user embedding for dynamic networks \cite{foulds2011dynamic,xu2014dynamic} and a dynamic extension of LDA \cite{blei2006dynamic}. 
Many variants of HMMs such as hidden semi-Markov models, autoregressive HMMs --- as well as their discriminative counterpart, Conditional random fields (CRF) ---
have been proposed to better model temporal structure. 
In the context of social media data, the stochastic block transition model \cite{xu2015stochastic} combined a dynamic discrete user embedding with an autoregressive HMM to model wall posts on Facebook over 90-day time bins.


\paragraph{Probabilistic models for continuous time}
Unlike discrete-time models, which cannot capture changes in behavior within a time bin, continuous-time models capture user behavior at arbitrary time points. 
This is useful when modeling user activities on social media, which are often quite bursty.  
A temporal point process (TPP) is a generative model for a sequence of times $(\Delta_1, \Delta_2, \ldots)$ between events, e.g.~posts on social media, from which one can obtain exact times of events $(t_1, t_2, \ldots)$ by summing inter-event times. 
In addition to being bursty, the distribution of inter-event times has also been found to be bimodal, which motivated the ACT-M model for temporal activity on social media \cite{costa2017modeling}. 

A common TPP used to model user activities on social media is the Hawkes process \cite{hawkes2018hawkes}, which is self-exciting so that the occurrence of an event increases the probability of another event occurring shortly thereafter. 
A multivariate Hawkes process models multiple time sequences that are mutually exciting. 
Multivariate Hawkes processes were combined with latent space models \cite{hoff2002latent} to learn user embeddings that decouple homophily and reciprocity from Facebook wall posts \cite{yang2017decoupling}. 
Marked TPPs augment TPPs with marks for events, which model additional information beyond event times, such as type of user action (e.g.~like or dislike), text and other content types. 
User-specific marked multivariate Hawkes processes were used by \cite{khodadadi2018continuous} to model questions and answers on Stack Overflow with marks denoting badges that users may earn.
Unlike discrete-time probabilistic models, the user embeddings in continuous-time models are usually static.

\paragraph{Deep learning for discrete time}
As previously stated, recurrent neural networks (RNN) are a common choice in modeling dynamic user embeddings from temporal structured data~\cite{kumar2019predicting,gong2020exploring,wu2017sequential}. The RNN (and its variants GRU, LSTM) take a sequence of vectors as input and apply a recurrence formula at every discrete time step --- same set of parameters are shared between these steps.
For example, ~\cite{kumar2019predicting} utilized RNN to project user and item embeddings at discrete time steps, eventually use the generated embeddings to predict future user--item interactions and also predict user state change (e.g. Reddit bans).
However, RNNs can only model the embeddings at discrete intervals even if the temporal data itself can be used as continuous input~\cite{kumar2019predicting} -- this makes it difficult to model the temporal dynamics of the embedding at any arbitrary timestamp.


\paragraph{Deep learning for continuous time}
%
%
As an alternative to RNN-based models, \cite{nguyen2018continuous} proposed a continuous-time dynamic network embedding approach based on temporal random walks. The embedding is computed by using sampled random walks that follow a time ordering and fit within a minimum and maximum time window. 
 
Recurrent Marked Temporal Point Process (RMTPP), a  generative model for event times that combines marked temporal point processes with RNNs is presented in \cite{du2016recurrent}. 
Unlikely the purely probabilistic TPP models, which typically specify a parametric model for the event times, the RMTPP uses an RNN to jointly model the timestamp and mark sequences over time.




\section{Current Limitation \& Future Directions }
We next identify several issues/limitations in the current research, which could be addressed in future research.

\noindent \textbf{Continuous-time deep learning models}
In social media data user activities (e.g., posting) rarely happen at regular intervals. 
Traditional auto-regressive models (RNN/LSTM) are limited to learning representations at discrete steps, making it difficult for them to generate representations at arbitrary time points. 
There is emerging work on continuous-time deep learning approaches~\cite{du2016recurrent,nguyen2018continuous}, but this area is relatively understudied. As a recent alternative which could be employed, 
Neural Ordinary Differential Equations (Neural ODE)~\cite{chen2018neural} are a new family of deep neural network models capable of building continuous-time series models. 
Neural ODE parameterizes the derivatives of hidden states using a neural network.

\noindent \textbf{Fair and ethical temporal user embedding} Recently, there is a surge of research interest in fair AI and machine learning. As demonstrated in a recent study~\cite{Islam2021}, user embeddings learned from social media data exhibit biases (e.g., gender and age bias). So far, there has not been much work on developing fair AI/ML techniques to ensure that the temporal user embeddings learned from social media data are unbiased and will not encode prejudice against marginalized groups. Moreover, protecting the privacy of social media users is of utmost importance. There is thus a need for privacy-preserving user embedding and analysis. 

\noindent \textbf{Generalizable time embedding}  Currently, except for Time2Vec~\cite{kazemi2019time2vec}, there has not been much work that produces a model-agnostic vector representation of time that can easily be incorporated into existing machine learning architectures. 

\noindent \textbf{Learning user embeddings with multimodal data} Most work in our survey uses only a single type of temporal user data (e.g., a sequence of text posts or a dynamic social network). Since there are diverse types of user data on social media, learning user embeddings from multimodal temporal data may allow us to build more comprehensive and more accurate user representations. 

\noindent \textbf{ Dynamic user embedding}  Most existing work learns a static embedding from temporal user data. As user interests and behavior may evolve over time,  the learned user representations should also vary with time. More work is needed to learn dynamic user embeddings that change with time. 

\noindent \textbf{Explainability} As user embeddings are latent feature vectors that can be difficult to interpret, there is an urgent need to develop new explainable AI technologies (e.g. visualization) to help users to gain insight into the embedding models.  

\noindent \textbf{Resource consciousness} Many of the state-of-the-art deep learning-based embedding models are very large (e.g., with billions of parameters ) and can be very expensive to train. For example,  BERT~\cite{devlin2018bert}, a neural network model that is frequently used to learn text embeddings requires high-performance GPU or TPU servers to train. This can prevent those who do not have expensive hardware and resources from trying these models. To make embedding technologies more accessible, there is an urgent need to develop novel techniques to build more concise and more efficient models that are resource conscious. 


\noindent \textbf{Conclusion} User embeddings enable the automated understanding of social media users, with implications for e-commerce, social science, and social good applications. We surveyed work addressing the essential temporal nature of social media. As we have seen, many strides have been taken, but much remains to be done in this important research area.

\bibliographystyle{named}
\bibliography{main}

\begin{thebibliography}{}

\bibitem[\protect\citeauthoryear{Amir \bgroup \em et al.\egroup
  }{2017}]{amir2017quantifying}
Silvio Amir, Glen Coppersmith, Paula Carvalho, M{\'a}rio~J Silva, and Bryon~C
  Wallace.
\newblock Quantifying mental health from social media with neural user
  embeddings.
\newblock In {\em MLHC}, pages 306--321. PMLR, 2017.

\bibitem[\protect\citeauthoryear{Beladev \bgroup \em et al.\egroup
  }{2020}]{beladev2020tdgraphembed}
Moran Beladev, Lior Rokach, Gilad Katz, Ido Guy, and Kira Radinsky.
\newblock tdgraphembed: Temporal dynamic graph-level embedding.
\newblock In {\em Proceedings of the 29th ACM International Conference on
  Information \& Knowledge Management}, pages 55--64, 2020.

\bibitem[\protect\citeauthoryear{Blei and Lafferty}{2006}]{blei2006dynamic}
David~M Blei and John~D Lafferty.
\newblock Dynamic topic models.
\newblock In {\em Proceedings of the 23rd international conference on Machine
  learning}, pages 113--120, 2006.

\bibitem[\protect\citeauthoryear{Blei \bgroup \em et al.\egroup
  }{2003}]{blei2003latent}
David~M Blei, Andrew~Y Ng, and Michael~I Jordan.
\newblock Latent dirichlet allocation.
\newblock {\em the Journal of machine Learning research}, 3:993--1022, 2003.

\bibitem[\protect\citeauthoryear{Chen \bgroup \em et al.\egroup
  }{2018}]{chen2018neural}
Ricky~TQ Chen, Yulia Rubanova, Jesse Bettencourt, and David Duvenaud.
\newblock Neural ordinary differential equations.
\newblock In {\em Proceedings of the 32nd International Conference on Neural
  Information Processing Systems}, pages 6572--6583, 2018.

\bibitem[\protect\citeauthoryear{Costa \bgroup \em et al.\egroup
  }{2017}]{costa2017modeling}
Alceu~Ferraz Costa, Yuto Yamaguchi, Agma Juci~Machado Traina, Caetano~Traina
  Jr, and Christos Faloutsos.
\newblock Modeling temporal activity to detect anomalous behavior in social
  media.
\newblock {\em ACM Transactions on Knowledge Discovery from Data (TKDD)},
  11(4):1--23, 2017.

\bibitem[\protect\citeauthoryear{Devlin \bgroup \em et al.\egroup
  }{2018}]{devlin2018bert}
Jacob Devlin, Ming-Wei Chang, Kenton Lee, and Kristina Toutanova.
\newblock Bert: Pre-training of deep bidirectional transformers for language
  understanding.
\newblock {\em arXiv:1810.04805}, 2018.

\bibitem[\protect\citeauthoryear{Diao \bgroup \em et al.\egroup
  }{2012}]{diao2012finding}
Qiming Diao, Jing Jiang, Feida Zhu, and Ee~Peng LIM.
\newblock Finding bursty topics from microblogs.
\newblock Acl, 2012.

\bibitem[\protect\citeauthoryear{Ding \bgroup \em et al.\egroup
  }{2017}]{ding2017multi}
Tao Ding, Warren~K Bickel, and Shimei Pan.
\newblock Multi-view unsupervised user feature embedding for social media-based
  substance use prediction.
\newblock In {\em Proceedings of the 2017 Conference on Empirical Methods in
  Natural Language Processing}, pages 2275--2284, 2017.

\bibitem[\protect\citeauthoryear{Ding \bgroup \em et al.\egroup
  }{2018}]{ding2018interpreting}
Tao Ding, Fatema Hasan, Warren~K Bickel, and Shimei Pan.
\newblock Interpreting social media-based substance use prediction models with
  knowledge distillation.
\newblock In {\em 2018 IEEE 30th International Conference on Tools with
  Artificial Intelligence (ICTAI)}, pages 623--630. IEEE, 2018.

\bibitem[\protect\citeauthoryear{Du \bgroup \em et al.\egroup
  }{2016}]{du2016recurrent}
Nan Du, Hanjun Dai, Rakshit Trivedi, Utkarsh Upadhyay, Manuel Gomez-Rodriguez,
  and Le~Song.
\newblock Recurrent marked temporal point processes: Embedding event history to
  vector.
\newblock In {\em KDD'16}, pages 1555--1564, 2016.

\bibitem[\protect\citeauthoryear{Fani \bgroup \em et al.\egroup
  }{2020}]{fani2020user}
Hossein Fani, Eric Jiang, Ebrahim Bagheri, Feras Al-Obeidat, Weichang Du, and
  Mehdi Kargar.
\newblock User community detection via embedding of social network structure
  and temporal content.
\newblock {\em Information Processing \& Management}, 57(2):102056, 2020.

\bibitem[\protect\citeauthoryear{Foulds \bgroup \em et al.\egroup
  }{2011}]{foulds2011dynamic}
James Foulds, Christopher DuBois, Arthur Asuncion, Carter Butts, and Padhraic
  Smyth.
\newblock A dynamic relational infinite feature model for longitudinal social
  networks.
\newblock In {\em AISTATS 2011}, pages 287--295. JMLR Workshop and Conference
  Proceedings, 2011.

\bibitem[\protect\citeauthoryear{Gao \bgroup \em et al.\egroup
  }{2017}]{gao2017novel}
Sheng Gao, Huacan Pang, Patrick Gallinari, Jun Guo, and Nei Kato.
\newblock A novel embedding method for information diffusion prediction in
  social network big data.
\newblock {\em IEEE Transactions on Industrial Informatics}, 13(4):2097--2105,
  2017.

\bibitem[\protect\citeauthoryear{Gong \bgroup \em et al.\egroup
  }{2020}]{gong2020exploring}
Maoguo Gong, Shunfei Ji, Yu~Xie, Yuan Gao, and AK~Qin.
\newblock Exploring temporal information for dynamic network embedding.
\newblock {\em IEEE Transactions on Knowledge and Data Engineering}, 2020.

\bibitem[\protect\citeauthoryear{Hawkes}{2018}]{hawkes2018hawkes}
Alan~G Hawkes.
\newblock Hawkes processes and their applications to finance: a review.
\newblock {\em Quantitative Finance}, 18(2):193--198, 2018.

\bibitem[\protect\citeauthoryear{Hoff \bgroup \em et al.\egroup
  }{2002}]{hoff2002latent}
Peter~D Hoff, Adrian~E Raftery, and Mark~S Handcock.
\newblock Latent space approaches to social network analysis.
\newblock {\em JASA}, 97(460):1090--1098, 2002.

\bibitem[\protect\citeauthoryear{Islam \bgroup \em et al.\egroup
  }{2021}]{Islam2021}
R.~Islam, K.N. Keya, Z.~Zeng, S.~Pan, and J.R. Foulds.
\newblock Debiasing career recommendations with neural fair collaborative
  filtering.
\newblock In {\em {WWW} '21}. {ACM} / {IW3C2}, 2021.

\bibitem[\protect\citeauthoryear{Kazemi \bgroup \em et al.\egroup
  }{2019}]{kazemi2019time2vec}
Seyed~Mehran Kazemi, Rishab Goel, Sepehr Eghbali, Janahan Ramanan, Jaspreet
  Sahota, Sanjay Thakur, Stella Wu, Cathal Smyth, Pascal Poupart, and Marcus
  Brubaker.
\newblock Time2vec: Learning a vector representation of time.
\newblock {\em arXiv:1907.05321}, 2019.

\bibitem[\protect\citeauthoryear{Khodadadi \bgroup \em et al.\egroup
  }{2018}]{khodadadi2018continuous}
Ali Khodadadi, Seyed~Abbas Hosseini, Erfan Tavakoli, and Hamid~R Rabiee.
\newblock Continuous-time user modeling in presence of badges: A probabilistic
  approach.
\newblock {\em ACM Transactions on Knowledge Discovery from Data (TKDD)},
  12(3):1--30, 2018.

\bibitem[\protect\citeauthoryear{Kosinski \bgroup \em et al.\egroup
  }{2013}]{kosinski2013private}
Michal Kosinski, David Stillwell, and Thore Graepel.
\newblock Private traits and attributes are predictable from digital records of
  human behavior.
\newblock {\em PNAS}, 110(15):5802--5805, 2013.

\bibitem[\protect\citeauthoryear{Kumar \bgroup \em et al.\egroup
  }{2019}]{kumar2019predicting}
Srijan Kumar, Xikun Zhang, and Jure Leskovec.
\newblock Predicting dynamic embedding trajectory in temporal interaction
  networks.
\newblock In {\em Proceedings of the 25th ACM SIGKDD International Conference
  on Knowledge Discovery \& Data Mining}, pages 1269--1278, 2019.

\bibitem[\protect\citeauthoryear{Li \bgroup \em et al.\egroup
  }{2017}]{li2017attributed}
Jundong Li, Harsh Dani, Xia Hu, Jiliang Tang, Yi~Chang, and Huan Liu.
\newblock Attributed network embedding for learning in a dynamic environment.
\newblock In {\em Proceedings of the 2017 ACM on Conference on Information and
  Knowledge Management}, pages 387--396, 2017.

\bibitem[\protect\citeauthoryear{Liang \bgroup \em et al.\egroup
  }{2018}]{liang2018dynamic}
Shangsong Liang, Xiangliang Zhang, Zhaochun Ren, and Evangelos Kanoulas.
\newblock Dynamic embeddings for user profiling in twitter.
\newblock In {\em Proceedings of the 24th ACM SIGKDD International Conference
  on Knowledge Discovery \& Data Mining}, pages 1764--1773, 2018.

\bibitem[\protect\citeauthoryear{Liu \bgroup \em et al.\egroup
  }{2020}]{liu2020dynamic}
Zhijun Liu, Chao Huang, Yanwei Yu, Peng Song, Baode Fan, and Junyu Dong.
\newblock Dynamic representation learning for large-scale attributed networks.
\newblock In {\em Proceedings of the 29th ACM International Conference on
  Information \& Knowledge Management}, pages 1005--1014, 2020.

\bibitem[\protect\citeauthoryear{Mikolov \bgroup \em et al.\egroup
  }{2013}]{mikolov2013distributed}
Tomas Mikolov, Ilya Sutskever, Kai Chen, Greg Corrado, and Jeffrey Dean.
\newblock Distributed representations of words and phrases and their
  compositionality.
\newblock In {\em Neurips 2013}, pages 3111--3119, 2013.

\bibitem[\protect\citeauthoryear{Nguyen \bgroup \em et al.\egroup
  }{2018}]{nguyen2018continuous}
Giang~Hoang Nguyen, John~Boaz Lee, Ryan~A Rossi, Nesreen~K Ahmed, Eunyee Koh,
  and Sungchul Kim.
\newblock Continuous-time dynamic network embeddings.
\newblock In {\em WWW'18: Companion Proceedings}, pages 969--976, 2018.

\bibitem[\protect\citeauthoryear{Noorshams \bgroup \em et al.\egroup
  }{2020}]{noorshams2020ties}
Nima Noorshams, Saurabh Verma, and Aude Hofleitner.
\newblock Ties: Temporal interaction embeddings for enhancing social media
  integrity at facebook.
\newblock In {\em Proceedings of the 26th ACM SIGKDD International Conference
  on Knowledge Discovery \& Data Mining}, pages 3128--3135, 2020.

\bibitem[\protect\citeauthoryear{Pan and Ding}{2019}]{pan2019social}
Shimei Pan and Tao Ding.
\newblock Social media-based user embedding: A literature review.
\newblock {\em arXiv preprint arXiv:1907.00725}, 2019.

\bibitem[\protect\citeauthoryear{Pennacchiotti and
  Popescu}{2011}]{pennacchiotti2011machine}
Marco Pennacchiotti and Ana-Maria Popescu.
\newblock A machine learning approach to twitter user classification.
\newblock In {\em ICWSM}, volume~5, 2011.

\bibitem[\protect\citeauthoryear{Qiu \bgroup \em et al.\egroup
  }{2020}]{qiu2020temporal}
Zhenyu Qiu, Wenbin Hu, Jia Wu, Weiwei Liu, Bo~Du, and Xiaohua Jia.
\newblock Temporal network embedding with high-order nonlinear information.
\newblock In {\em Proceedings of the AAAI Conference on Artificial
  Intelligence}, volume~34, pages 5436--5443, 2020.

\bibitem[\protect\citeauthoryear{Sang \bgroup \em et al.\egroup
  }{2015}]{sang2015probabilistic}
Jitao Sang, Dongyuan Lu, and Changsheng Xu.
\newblock A probabilistic framework for temporal user modeling on microblogs.
\newblock In {\em Proceedings of the 24th ACM International on Conference on
  Information and Knowledge Management}, pages 961--970, 2015.

\bibitem[\protect\citeauthoryear{Wu \bgroup \em et al.\egroup
  }{2016}]{wu2016unfolding}
Bo~Wu, Tao Mei, Wen-Huang Cheng, and Yongdong Zhang.
\newblock Unfolding temporal dynamics: Predicting social media popularity using
  multi-scale temporal decomposition.
\newblock In {\em Proceedings of the AAAI Conference on Artificial
  Intelligence}, volume~30, 2016.

\bibitem[\protect\citeauthoryear{Wu \bgroup \em et al.\egroup
  }{2017}]{wu2017sequential}
Bo~Wu, Wen-Huang Cheng, Yongdong Zhang, Qiushi Huang, Jintao Li, and Tao Mei.
\newblock Sequential prediction of social media popularity with deep temporal
  context networks.
\newblock {\em arXiv preprint arXiv:1712.04443}, 2017.

\bibitem[\protect\citeauthoryear{Xie \bgroup \em et al.\egroup
  }{2016}]{xie2016learning}
Min Xie, Hongzhi Yin, Hao Wang, Fanjiang Xu, Weitong Chen, and Sen Wang.
\newblock Learning graph-based poi embedding for location-based recommendation.
\newblock In {\em Proceedings of the 25th ACM International on Conference on
  Information and Knowledge Management}, pages 15--24, 2016.

\bibitem[\protect\citeauthoryear{Xiong \bgroup \em et al.\egroup
  }{2019}]{xiong2019dyngraphgan}
Yun Xiong, Yao Zhang, Hanjie Fu, Wei Wang, Yangyong Zhu, and S~Yu Philip.
\newblock Dyngraphgan: Dynamic graph embedding via generative adversarial
  networks.
\newblock In {\em International Conference on Database Systems for Advanced
  Applications}, pages 536--552. Springer, 2019.

\bibitem[\protect\citeauthoryear{Xu and Hero}{2014}]{xu2014dynamic}
Kevin~S Xu and Alfred~O Hero.
\newblock Dynamic stochastic blockmodels for time-evolving social networks.
\newblock {\em IEEE Journal of Selected Topics in Signal Processing},
  8(4):552--562, 2014.

\bibitem[\protect\citeauthoryear{Xu}{2015}]{xu2015stochastic}
Kevin Xu.
\newblock Stochastic block transition models for dynamic networks.
\newblock In {\em AISTATS}, pages 1079--1087. PMLR, 2015.

\bibitem[\protect\citeauthoryear{Yang \bgroup \em et al.\egroup
  }{2017}]{yang2017decoupling}
Jiasen Yang, Vinayak~A Rao, and Jennifer Neville.
\newblock Decoupling homophily and reciprocity with latent space network
  models.
\newblock In {\em UAI}, 2017.

\bibitem[\protect\citeauthoryear{Yin \bgroup \em et al.\egroup
  }{2013}]{yin2013unified}
Hongzhi Yin, Bin Cui, Hua Lu, Yuxin Huang, and Junjie Yao.
\newblock A unified model for stable and temporal topic detection from social
  media data.
\newblock In {\em 2013 IEEE 29th International Conference on Data Engineering
  (ICDE)}, pages 661--672. IEEE, 2013.

\bibitem[\protect\citeauthoryear{Yin \bgroup \em et al.\egroup
  }{2014}]{yin2014temporal}
Hongzhi Yin, Bin Cui, Ling Chen, Zhiting Hu, and Zi~Huang.
\newblock A temporal context-aware model for user behavior modeling in social
  media systems.
\newblock In {\em Proceedings of the 2014 ACM SIGMOD international conference
  on Management of data}, pages 1543--1554, 2014.

\bibitem[\protect\citeauthoryear{Yin \bgroup \em et al.\egroup
  }{2015}]{yin2015dynamic}
Hongzhi Yin, Bin Cui, Ling Chen, Zhiting Hu, and Xiaofang Zhou.
\newblock Dynamic user modeling in social media systems.
\newblock {\em ACM Transactions on Information Systems (TOIS)}, 33(3):1--44,
  2015.

\bibitem[\protect\citeauthoryear{Yu \bgroup \em et al.\egroup
  }{2018}]{yu2018netwalk}
Wenchao Yu, Wei Cheng, Charu~C Aggarwal, Kai Zhang, Haifeng Chen, and Wei Wang.
\newblock Netwalk: A flexible deep embedding approach for anomaly detection in
  dynamic networks.
\newblock In {\em Proceedings of the 24th ACM SIGKDD International Conference
  on Knowledge Discovery \& Data Mining}, pages 2672--2681, 2018.

\bibitem[\protect\citeauthoryear{Zhang \bgroup \em et al.\egroup
  }{2018}]{zhang2018user}
Wei Zhang, Wen Wang, Jun Wang, and Hongyuan Zha.
\newblock User-guided hierarchical attention network for multi-modal social
  image popularity prediction.
\newblock In {\em WWW 2018}, pages 1277--1286, 2018.

\bibitem[\protect\citeauthoryear{Zhao \bgroup \em et al.\egroup
  }{2017}]{zhao2017geo}
Shenglin Zhao, Tong Zhao, Irwin King, and Michael~R Lyu.
\newblock Geo-teaser: Geo-temporal sequential embedding rank for
  point-of-interest recommendation.
\newblock In {\em Proceedings of the 26th international conference on world
  wide web companion}, pages 153--162, 2017.

\bibitem[\protect\citeauthoryear{Zhou \bgroup \em et al.\egroup
  }{2018}]{zhou2018dynamic}
Lekui Zhou, Yang Yang, Xiang Ren, Fei Wu, and Yueting Zhuang.
\newblock Dynamic network embedding by modeling triadic closure process.
\newblock In {\em AAAI}, volume~32, 2018.

\bibitem[\protect\citeauthoryear{Zhuo \bgroup \em et al.\egroup
  }{2019}]{zhuo2019diffusiongan}
Wei Zhuo, Yanan Zhao, Qianyi Zhan, and Yuan Liu.
\newblock Diffusiongan: Network embedding for information diffusion prediction
  with generative adversarial nets.
\newblock In {\em ISPA/BDCloud/SocialCom/SustainCom}, pages 808--816. IEEE,
  2019.

\end{thebibliography}
\end{document}